\documentclass[runningheads]{llncs}
\usepackage[T1]{fontenc}
\usepackage{indentfirst}
\usepackage[colorlinks=true, urlcolor=blue, linkcolor=red]{hyperref}
\usepackage{graphicx}
\usepackage{amsmath}
\usepackage[section]{placeins}
\usepackage{booktabs}
\usepackage{cleveref}
\begin{document}

\title{Divide and Conquer: Grounding a Bleeding Areas in Gastrointestinal Image with Two-Stage Model}

%
%
\author{Yu-Fan Lin \and Bo-Cheng Qiu \and Chia-Ming Lee \and Chih-Chung Hsu  \thanks{means the corresponding author.}}
\authorrunning{Y.F. Lin et al.}
%
\institute{Institute of Data Science, National Cheng Kung University, Taiwan\\
\email{\{aas12as12as12tw, a36492183, zuw408421476\}@gmail.com}\\
\email{\inst{*}cchsu@gs.ncku.edu.tw}}
\maketitle              
\begin{abstract}
Accurate detection and segmentation of gastrointestinal bleeding are critical for diagnosing diseases such as peptic ulcers and colorectal cancer. This study proposes a two-stage framework that decouples classification and grounding to address the inherent challenges posed by traditional Multi-Task Learning models, which jointly optimizes classification and segmentation. Our approach separates these tasks to achieve targeted optimization for each. The model first classifies images as bleeding or non-bleeding, thereby isolating subsequent grounding from inter-task interference and label heterogeneity. To further enhance performance, we incorporate Stochastic Weight Averaging and Test-Time Augmentation, which improve model robustness against domain shifts and annotation inconsistencies. Our method is validated on the Auto-WCEBleedGen Challenge V2 Challenge dataset and achieving second place. Experimental results demonstrate significant improvements in classification accuracy and segmentation precision, especially on sequential datasets with consistent visual patterns. This study highlights the practical benefits of a two-stage strategy for medical image analysis and sets a new standard for GI bleeding detection and segmentation. Our code is publicly available at \href{https://github.com/VanLinLin/Auto-WCEBleedGen-Challenge-Version-V2/tree/main}{this GitHub repository}.

\end{abstract}

\section{Introduction}
With the development of deep learning, artificial intelligence has increasingly assisted in automatic medical diagnosis to reduce labor costs. Applications such as bleed detection \cite{hub2024auto,WCEbleedGen,AutoWCEB(Improved)} and COVID-19 detection \cite{10192945,10678126,hsu2023strongbaselinebagtricks,hsu2024simple2dconvolutionalneural} exemplify its growing impact. Gastrointestinal (GI) bleeding, a critical indicator for diagnosing digestive system diseases like peptic ulcers and colorectal cancer, highlights the importance of detecting and localizing bleeding regions within GI images. Accurate classification and grounding enable early detection, precise diagnosis, and targeted treatment. With the increasing use of Wireless Capsule Endoscopy (WCE) for GI imaging, there is a strong demand for automated models capable of both classification and grounding to support real-time clinical decisions.

Multi-task learning (MTL) frameworks have been widely applied across various fields, such as autonomous driving \cite{feng2020deep}, natural language processing \cite{shuang2019adversarial}, and medical imaging \cite{chowdary2022multi}. These frameworks leverage shared representations to achieve multiple objectives simultaneously, offering efficiency in model size and computational cost \cite{crawshaw2020multi}. However, applying MTL to GI bleeding detection introduces significant challenges:

\begin{itemize} \item \textbf{Imbalanced class distribution}: The uneven representation of bleeding and non-bleeding images skews the model's learning process, often leading to biased predictions favoring the majority class. \item \textbf{Heterogeneous data sources}: GI endoscopic images capture varying regions within the digestive tract, contributing to significant heterogeneity due to patient differences, imaging devices, and bleeding locations. This exacerbates domain shifts and increases the difficulty of generalizing across diverse datasets. \item \textbf{Annotation inconsistencies}: Inconsistent or subjective annotations result in noisy labels, increasing training complexity and reducing model reliability. \item \textbf{Scarcity of medical samples}: The limited availability of annotated medical datasets further restricts the ability to train robust models capable of handling real-world variability. \end{itemize}

Our two-stage framework addresses these challenges by decoupling the tasks of classification and grounding. In the first stage, images are classified to identify those containing bleeding, effectively filtering out non-bleeding images. This step reduces the heterogeneity of data passed to the grounding stage, where more focused and effective detection and segmentation are performed. By separating these tasks, our approach mitigates inter-task conflicts common in MTL models and enables independent optimization for each stage.

To further enhance performance, we incorporate Stochastic Weight Averaging (SWA) \cite{2} and Test-Time Augmentation (TTA) \cite{1}. SWA stabilizes model convergence and enhances generalization by averaging weights over training iterations, while TTA applies multiple test-time transformations to improve prediction robustness. These techniques help address domain shifts, annotation inconsistencies, and data scarcity, ultimately boosting model performance.

In summary, this study introduces a two-stage framework for the classification and grounding of bleeding in WCE images. By decoupling tasks, addressing dataset-specific challenges, and integrating advanced techniques like SWA and TTA, the framework achieves superior performance and robustness. These advancements enabled the framework to secure second place in the Auto-WCEBleedGen Challenge V2. This two-stage strategy demonstrates practical value in medical image analysis and holds promise for diverse clinical applications, paving the way for improved detection and segmentation in gastrointestinal endoscopy.

\section{Methodolodgy}
\subsection{Model architecture} 
\begin{figure*}[h]
    \centering
    \includegraphics[width=1\linewidth]{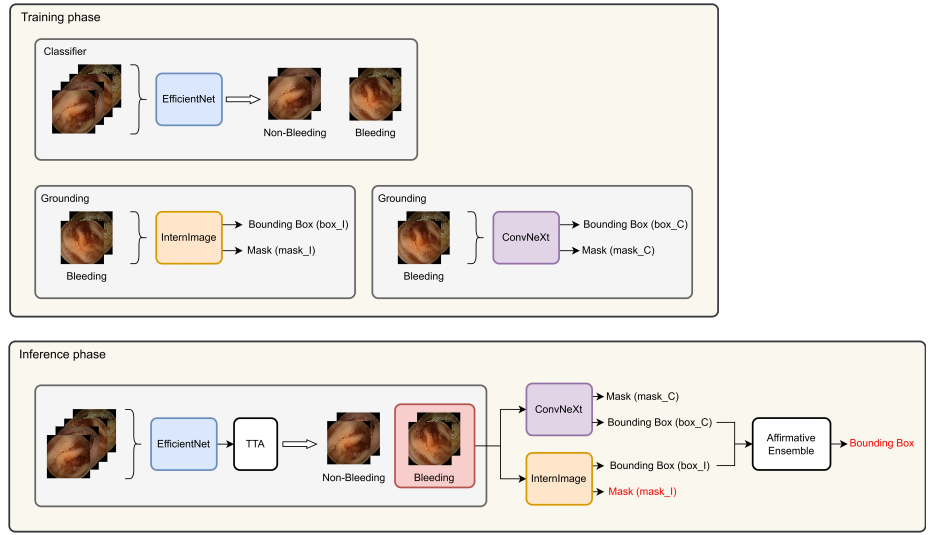}
    \caption{Our proposed method consists of a classification model and two instance segmentation models during the training phase (top figure), which are combined during the inference phase (bottom figure). In the training stage, the classification model is trained to differentiate between bleeding and non-bleeding images, with the identified bleeding images used to train the two instance segmentation models. SWA is applied during the final ten epochs of training for each model to improve model stability and generalization. During the inference phase, the classification model is enhanced with TTA and sequentially integrated with the instance segmentation models to form an optimized pipeline for robust performance. The red text indicates the final adopted output.}
    \label{fig:model1}
\end{figure*}

The model construction is divided into two stages: classification and segmentation. As illustrated in Figure \ref{fig:model1}, beginning with classification, data augmentation is performed first, employing the EfficientNet-B7 \cite{4,rw2019timm} for categorization. During this phase, SWA \cite{2} and TTA \cite{1} are incorporated to bolster the model's generalizability and performance. Once intestinal images with bleeding are classified, another round of data augmentation is applied. The bleeding images are then processed using both the ConvNeXt \cite{6} and InternImage \cite{7}. 

The choice of ConvNeXt and InternImage as feature extractors is based on their superior performance in capturing rich and hierarchical visual features. ConvNeXt is a modernized version of convolutional networks that leverages design principles from vision transformers, offering a balance between efficiency and accuracy. InternImage, on the other hand, is designed for high-resolution imagery, excelling in dense prediction tasks with its dynamic and adaptive convolutional mechanisms. These models provide robust feature representation, making them well-suited for tasks requiring detailed medical image understanding.

Afterwards, we apply SWA and TTA being utilized again to enhance the results. Finally, to improve the predictions of the bounding boxes, an affirmative strategy is adopted to ensemble ConvNeXt and InternImage, aiming to refine the detection accuracy for final results.


\begin{figure*}[h]
    \centering
    \includegraphics[width=0.8\linewidth]{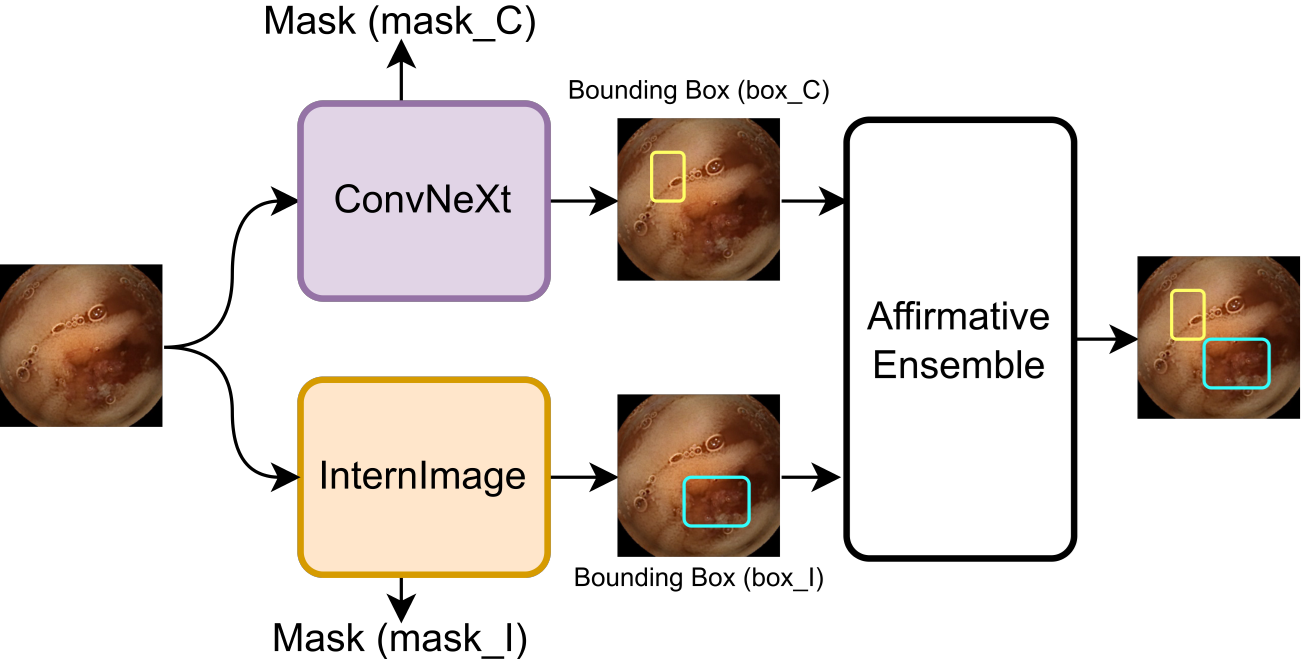}
    \caption{The illustration of affirmative ensemble.}
    \label{fig:model}
\end{figure*}

\subsection{Post-processing}
\textbf{Stochastic Weight Averaging.} SWA is a technique designed to enhance the generalization performance of models by averaging their weights \cite{2}. The fundamental concept of SWA lies in utilizing multiple sets of model weights collected during the training process and computing their average to derive the final model weights. We applied the SWA technique during the final 10 epochs of training for both the classification and grounding models. By averaging the converged weights, we obtained more stable and generalizable final weights.

\textbf{Testing Time Augmentation.} TTA is a technique aimed at enhancing model performance during the inference stage \cite{1}. The core idea of TTA involves applying various data augmentation methods to the test data and aggregating the model's predictions on all augmented samples, such as by averaging the predictions or using majority voting. This process leads to more stable and accurate final predictions.

\textbf{Affirmative Ensemble.} Affirmative Ensemble for bounding box is a specialized ensemble method tailored for bounding box prediction, designed to enhance the accuracy and stability of object detection models. The core concept of this approach lies in aggregating predictions from multiple models to generate more reliable and precise final bounding boxes, thereby effectively reducing the uncertainty and errors associated with individual model predictions. We applied this technique to the integration of bleeding region detection, adopting a strategy that treats all bounding boxes as valid candidates for ensemble processing. Although this approach increases the false positive rate, it concurrently improves the sensitivity to suspicious regions. In the medical domain, we consider this trade-off to be acceptable.

\section{Experiment}
\subsection{Experiments Setup}
The Auto-WCEBleedGen
Challenge V2 Challenge dataset comprises 2618 images, which is provided by MISAHUB \cite{hub2024auto,WCEbleedGen,AutoWCEB(Improved)}. We partitioned it into a training and validation set with an 8:2 ratio. Subsequently, the testing set consists of two distinct parts, comprising 49 and 515 frames, respectively. It is worth noting that the former consists of non-sequential images collected from different patients' organs, while the latter is a patient's gastro-oesophageal film record. We employed the MMDetection and MMPretrain \cite{chen2019mmdetection} frameworks to train all our models using two NVIDIA-GeForce RTX 3090 GPUs. The setup included CUDA version 11.7 and PyTorch version 1.13.0, along with Python 3.9.18.

\subsection{Hyper-parameter Settings}
Our pipeline is divided into two parts: classification and grounding. It shares some hyper-parameter settings. AdamW \cite{loshchilov2019decoupled} and Automatic Mixed Precision (AMP) \cite{micikevicius2018mixed} are used to optimize all models. The learning rate scheduler is StepLR.  All images are resized to 224 $\times$ 224 and undergo standard and common geometric and color-based augmentations, including rotation, flipping, RandAugment \cite{8}, and color jittering during training time.

\textbf{Classification Model.} For the classification task, we utilized EfficientNet-b7 \cite{4}, a state-of-the-art lightweight convolutional neural network known for its excellent performance and efficiency. The model was trained for $100$ epochs with a learning rate of $1.25e-4$. To optimize the model, we employed Cross-Entropy Loss as the objective function, which is widely used for multi-class classification problems. EfficientNet-b7's scalable architecture helped balance accuracy and computational efficiency, making it well-suited for this task.

\textbf{Grounding Model.} For grounding tasks, we leveraged ConvNeXt-Base \cite{6} and InternImage-XL \cite{7}, both of which are highly effective feature extractors for dense prediction tasks. These models were trained for $2000$ epochs with a learning rate of $1e-4$ and a weight decay of $5e-2$ to mitigate overfitting. For object detection, the loss function comprised the sum of Cross-Entropy Loss for classification and L1 Loss for bounding box regression, ensuring accurate localization and categorization of objects. For semantic segmentation, we employed a weighted combination of Cross-Entropy Loss and Dice Loss, effectively handling imbalanced classes and enhancing the segmentation quality. The choice of these advanced architectures ensured robust performance on complex grounding tasks.

After the training stage, we applied SWA for self-ensemble and combined it with TTA and the affirmative ensemble strategy to obtain the final results, as shown in the Figure \ref{fig:model}.

\subsection{Experiment Results}
In summary, we utilize a two-stage framework, which addresses the issue of distribution shifting in classification and grounding tasks. Subsequently, we enhanced the prediction results through self-ensemble and model ensemble techniques, as shown in \Cref{tab:result1,tab:result2}.

\textbf{Classification Results.} We summarize the classification results as shown in \Cref{tab:result1} for the two testing datasets. The disparity between the two datasets reflects the inherent differences in data distributions. Testing Data 1, which contains non-sequential images from various patients, poses challenges due to its higher heterogeneity. The diversity in visual features, such as lighting, anatomy, and bleeding patterns, introduces significant complexity in classification. In contrast, Testing Data 2, derived from a continuous sequence from a single patient, benefits from consistent visual and contextual patterns, enabling more robust classification results. These findings highlight the importance of data consistency and suggest that the proposed two-stage framework is particularly effective when data variability is controlled. The performance gap between the two datasets emphasizes the sensitivity of the model to dataset heterogeneity, underscoring the need for robust model design.

\textbf{Grounding Results.} As shown in \Cref{tab:result2}, the results for detection and segmentation reveal substantial performance disparities between the two datasets. For detection, the model’s ability to localize bleeding regions is significantly better on Testing Data 2 than on Testing Data 1, as evidenced by higher Average Precision (AP) and Average Recall (AR) metrics. This outcome highlights the model’s capability to leverage the structured nature of Testing Data 2, where sequential images provide consistent spatial and temporal information. Similarly, for segmentation, the AP and AR scores follow a similar trend, with Testing Data 2 significantly outperforming Testing Data 1 across all metrics. These results indicate that while the proposed two-stage framework can segment bleeding areas, its performance is influenced by the characteristics of the input data. The consistent structure of Testing Data 2 provides more stable spatial and temporal cues, thereby enhancing segmentation accuracy. By comparison, the heterogeneity of Testing Data 1, which consists of non-sequential images from multiple patients, results in greater variability in visual features, impacting the model’s grounding performance.

\begin{table}[!ht]
    \centering
    \caption{The results of classification on the testing set.}
    \begin{tabular}{l c c} 
      \toprule
                                    &Testing Data 1&Testing Data 2\\
      \midrule
      Accuracy                      &51.0204&74.9514\\
      \midrule
      Precision                     &50.0000&74.5802\\
      \midrule
      Recall                        &25.5102&74.7701\\
      \midrule
      F1-score                      &33.7837&74.6517\\
      \midrule
      \bottomrule
    \end{tabular}
    \label{tab:result1}
\end{table}

\begin{table}[!ht]
    \centering
    \caption{The results of grounding on the testing set.}
    \begin{tabular}{l c c c c} 
      \toprule                      
      Task & \multicolumn{2}{c}{detection} & \multicolumn{2}{c}{segmentation}\\
                &Testing1 &Testing2 &Testing1 &Testing2\\
      \midrule
      mAP@0.5:0.95         &15.8 &43.4 &6.50 &21.1\\
      \midrule
      mAP@0.5              &41.6 &72.3 &30.8 &51.4\\
      \midrule
      AP@0.5:0.95          &33.2 &56.5 &13.5 &29.3\\
      \midrule
      AP@0.5               &78.4 &86.2 &45.9 &61.2\\
      \midrule
      AR@0.5:0.95          &22.4 &42.1 &14.8 &25.2\\
      \midrule
      AR@0.5               &48.0 &61.2 &56.0 &52.5\\
      \bottomrule
    \end{tabular}
    \label{tab:result2}
\end{table}

\begin{figure}
    \centering
    \includegraphics[width=0.5\textwidth]{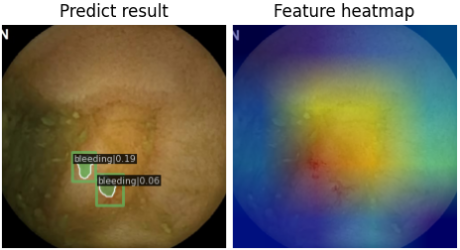}
    \caption{The Eigen-CAM \cite{jacobgilpytorchcam} visualization of the bleeding area.} 
    \label{featuremapintensity3.png}
\end{figure}

\textbf{Visualization Results.} 
The visualization of bleeding areas with Eigen-CAM \cite{jacobgilpytorchcam}, as shown in \Cref{featuremapintensity3.png}, illustrates the interpretability of the proposed framework. The heatmaps align well with the predicted bounding boxes, indicating that the model effectively focuses on clinically relevant regions during both detection and segmentation tasks. This alignment supports the efficacy of the two-stage framework in separating classification and segmentation tasks, thereby improving interpretability and reliability in medical imaging applications. 

\section{Conclusion}
In this report, we proposed the two-stage framework for grounding  Gastrointestinal intestinal bleeding region. The key idea of our method is to decouple three-different tasks in one stage into two-stage with various ensemble methods, including Testing-Time Augmentation and Stochastic Weight Averaging for self-ensemble, employing ConvNeXt and InternImage for model-ensemble. The framework achieved second place in the Auto-WCEBleedGen Challenge V2. Experiment results have demonstrated the proposed two-stage framework holds significant potential for improving clinical applications and setting a new standard in gastrointestinal bleeding detection and segmentation.

\section{Acknowledgments}
As participants in the Auto-WCEBleedGen Version V2 Challenge, we fully comply with the competition rules as outlined in \cite{hub2024auto} and the challenge website. Our methods have used the training and test data sets provided in the official release in \cite{WCEbleedGen} and \cite{AutoWCEB(Improved)} to report the results of the challenge.

\bibliographystyle{unsrt}
\bibliography{refs}

\begin{thebibliography}{10}

\bibitem{hub2024auto}
Palak Handa, Divyansh Nautiyal, Deepti Chhabra, Manas Dhir, Anushka Saini, Shreshtha Jha, Harshita Mangotra, Nishu Pandey, Advika Thakur, et~al.
\newblock Auto-wcebleedgen version v1 and v2: Challenge, datasets and evaluation.
\newblock {\em Authorea Preprints}, 2024.

\bibitem{WCEbleedGen}
Palak Handa, Manas Dhir, Amirreza Mahbod, Florian Schwarzhans, Ramona Woitek, Nidhi Goel, and Deepak Gunjan.
\newblock Wcebleedgen: A wireless capsule endoscopy dataset containing bleeding and non-bleeding frames.
\newblock \url{https://zenodo.org/records/10156571}, 2023.

\bibitem{AutoWCEB(Improved)}
Palak Handa, Nishu Pandey, Divyansh Nautiyal, Nidhi Goel, and Deepak Gunjan.
\newblock Autowcebleedgen-test dataset (improved).
\newblock \url{https://zenodo.org/records/10642779}, 2024.

\bibitem{10192945}
Chih-Chung Hsu, Chih-Yu Jian, Chia-Ming Lee, Chi-Han Tsai, and Shen-Chieh Tai.
\newblock Bag of tricks of hybrid network for covid-19 detection of ct scans.
\newblock In {\em 2023 IEEE International Conference on Acoustics, Speech, and Signal Processing Workshops (ICASSPW)}, pages 1--4, 2023.

\bibitem{10678126}
Chih-Chung Hsu, Chia-Ming Lee, Yang~Fan Chiang, Yi-Shiuan Chou, Chih-Yu Jiang, Shen-Chieh Tai, and Chi-Han Tsai.
\newblock A closer look at spatial-slice features learning for covid-19 detection.
\newblock In {\em 2024 IEEE/CVF Conference on Computer Vision and Pattern Recognition Workshops (CVPRW)}, pages 4924--4934, 2024.

\bibitem{hsu2023strongbaselinebagtricks}
Chih-Chung Hsu, Chih-Yu Jian, Chia-Ming Lee, Chi-Han Tsai, and Sheng-Chieh Dai.
\newblock Strong baseline and bag of tricks for covid-19 detection of ct scans, 2023.

\bibitem{hsu2024simple2dconvolutionalneural}
Chih-Chung Hsu, Chia-Ming Lee, Yang~Fan Chiang, Yi-Shiuan Chou, Chih-Yu Jiang, Shen-Chieh Tai, and Chi-Han Tsai.
\newblock Simple 2d convolutional neural network-based approach for covid-19 detection, 2024.

\bibitem{feng2020deep}
Di~Feng, Christian Haase-Sch{\"u}tz, Lars Rosenbaum, Heinz Hertlein, Claudius Glaeser, Fabian Timm, Werner Wiesbeck, and Klaus Dietmayer.
\newblock Deep multi-modal object detection and semantic segmentation for autonomous driving: Datasets, methods, and challenges.
\newblock {\em IEEE Transactions on Intelligent Transportation Systems}, 22(3):1341--1360, 2020.

\bibitem{shuang2019adversarial}
Kai Shuang, Meng Xu, Wentao Zhang, and Zhixuan Zhang.
\newblock Adversarial multi-task label embedding for text classification.
\newblock In {\em Proceedings of the 2019 2nd International Conference on Computational Intelligence and Intelligent Systems}, pages 45--50, 2019.

\bibitem{chowdary2022multi}
Jignesh Chowdary, Pratheepan Yogarajah, Priyanka Chaurasia, and Velmathi Guruviah.
\newblock A multi-task learning framework for automated segmentation and classification of breast tumors from ultrasound images.
\newblock {\em Ultrasonic imaging}, 44(1):3--12, 2022.

\bibitem{crawshaw2020multi}
Michael Crawshaw.
\newblock Multi-task learning with deep neural networks: A survey.
\newblock {\em arXiv preprint arXiv:2009.09796}, 2020.

\bibitem{2}
Pavel Izmailov, Dmitrii Podoprikhin, Timur Garipov, Dmitry Vetrov, and Andrew~Gordon Wilson.
\newblock Averaging weights leads to wider optima and better generalization.
\newblock {\em arXiv preprint arXiv:1803.05407}, 2018.

\bibitem{1}
Marvin Zhang, Sergey Levine, and Chelsea Finn.
\newblock Memo: Test time robustness via adaptation and augmentation.
\newblock {\em Advances in Neural Information Processing Systems}, 35:38629--38642, 2022.

\bibitem{4}
Mingxing Tan and Quoc Le.
\newblock Efficientnet: Rethinking model scaling for convolutional neural networks.
\newblock In {\em International conference on machine learning}, pages 6105--6114. PMLR, 2019.

\bibitem{rw2019timm}
Ross Wightman.
\newblock Pytorch image models.
\newblock \url{https://github.com/rwightman/pytorch-image-models}, 2019.

\bibitem{6}
Zhuang Liu, Hanzi Mao, Chao-Yuan Wu, Christoph Feichtenhofer, Trevor Darrell, and Saining Xie.
\newblock A convnet for the 2020s.
\newblock In {\em Proceedings of the IEEE/CVF Conference on Computer Vision and Pattern Recognition (CVPR)}, pages 11976--11986, June 2022.

\bibitem{7}
Wenhai Wang, Jifeng Dai, Zhe Chen, Zhenhang Huang, Zhiqi Li, Xizhou Zhu, Xiaowei Hu, Tong Lu, Lewei Lu, Hongsheng Li, Xiaogang Wang, and Yu~Qiao.
\newblock Internimage: Exploring large-scale vision foundation models with deformable convolutions.
\newblock In {\em Proceedings of the IEEE/CVF Conference on Computer Vision and Pattern Recognition (CVPR)}, pages 14408--14419, June 2023.

\bibitem{chen2019mmdetection}
Kai Chen, Jiaqi Wang, Jiangmiao Pang, Yuhang Cao, Yu~Xiong, Xiaoxiao Li, Shuyang Sun, Wansen Feng, Ziwei Liu, Jiarui Xu, Zheng Zhang, Dazhi Cheng, Chenchen Zhu, Tianheng Cheng, Qijie Zhao, Buyu Li, Xin Lu, Rui Zhu, Yue Wu, Jifeng Dai, Jingdong Wang, Jianping Shi, Wanli Ouyang, Chen~Change Loy, and Dahua Lin.
\newblock Mmdetection: Open mmlab detection toolbox and benchmark, 2019.

\bibitem{loshchilov2019decoupled}
Ilya Loshchilov and Frank Hutter.
\newblock Decoupled weight decay regularization, 2019.

\bibitem{micikevicius2018mixed}
Paulius Micikevicius, Sharan Narang, Jonah Alben, Gregory Diamos, Erich Elsen, David Garcia, Boris Ginsburg, Michael Houston, Oleksii Kuchaiev, Ganesh Venkatesh, and Hao Wu.
\newblock Mixed precision training, 2018.

\bibitem{8}
Ekin~D. Cubuk, Barret Zoph, Jonathon Shlens, and Quoc~V. Le.
\newblock Randaugment: Practical automated data augmentation with a reduced search space.
\newblock In {\em Proceedings of the IEEE/CVF Conference on Computer Vision and Pattern Recognition (CVPR) Workshops}, June 2020.

\bibitem{jacobgilpytorchcam}
Jacob Gildenblat and contributors.
\newblock Pytorch library for cam methods.
\newblock \url{https://github.com/jacobgil/pytorch-grad-cam}, 2021.

\end{thebibliography}
\end{document}